\pdfoutput=1

\documentclass[11pt]{article}

\usepackage[final]{acl}

\usepackage{times}
\usepackage{latexsym}

\usepackage[T1]{fontenc}

\usepackage[utf8]{inputenc}

\usepackage{microtype}

\usepackage{inconsolata}

\usepackage{graphicx}

\usepackage{multirow}
\usepackage{subfigure}

\usepackage{dsfont}
\usepackage{amssymb}
\usepackage{amsmath}
\usepackage{algorithm}
\usepackage{algorithmic}
\usepackage[skip=2.5pt]{caption}
\usepackage{array}
\usepackage{booktabs}

\title{
    A Semi-supervised Scalable Unified Framework for E-commerce Query Classification
}

\author{
 \textbf{Chunyuan Yuan\textsuperscript{1}},
 \textbf{Chong Zhang\textsuperscript{1}},
 \textbf{Zheng Fang\textsuperscript{1}},
 \textbf{Ming Pang\textsuperscript{1,*}},
\\
 \textbf{Xue Jiang\textsuperscript{1}},
 \textbf{Changping Peng\textsuperscript{1}},
 \textbf{Zhangang Lin\textsuperscript{1}},
 \textbf{Ching Law\textsuperscript{1}}
\\
 \textsuperscript{1}JD.COM,
\\
\small{
\texttt{\{yuanchunyuan1,zhangchong78,fangzheng21,pangming8,jiangxue,pengchangping,linzhangang,lawching\}@jd.com}
}
}

\begin{document}
\maketitle

\begin{abstract}
Query\let\thefootnote\relax\footnotetext{* Corresponding author.} classification, including multiple subtasks such as intent and category prediction, is vital to e-commerce applications. E-commerce queries are usually short and lack context, and the information between labels cannot be used, resulting in insufficient prior information for modeling. Most existing industrial query classification methods rely on users' posterior click behavior to construct training samples, resulting in a Matthew vicious cycle. Furthermore, the subtasks of query classification lack a unified framework, leading to low efficiency for algorithm optimization.

In this paper, we propose a novel \textbf{S}emi-supervised \textbf{S}calable \textbf{U}nified \textbf{F}ramework (SSUF), containing multiple enhanced modules to unify the query classification tasks. The knowledge-enhanced module uses world knowledge to enhance query representations and solve the problem of insufficient query information. The label-enhanced module uses label semantics and semi-supervised signals to reduce the dependence on posterior labels. The structure-enhanced module enhances the label representation based on the complex label relations. Each module is highly pluggable, and input features can be added or removed as needed according to each subtask. We conduct extensive offline and online A/B experiments, and the results show that SSUF significantly outperforms the state-of-the-art models.
\end{abstract}

\section{Introduction}
E-commerce platforms like Amazon, Taobao, and JD provide users with billions of diverse products and have become essential in our daily lives. Due to the wide variety of user needs and product categories, capturing users’ purchasing intentions is vital for both user experience and platform efficiency. Query classification, including intent, category, and brand prediction, plays a key role in understanding users’ shopping needs and supports the subsequent modules of the search system.

The inherent characteristics of e-commerce queries, which are typically short, and ambiguous, bring significant challenges for query classification. To solve the problem of insufficient information caused by short queries, some deep learning-based models, such as XML-CNN~\cite{liu2017deep}, KRF~\cite{ma2020beyond}, HiAGM~\cite{zhou2020hierarchy}, and LSAN~\cite{xiao2019label} have been proposed to learn the contextual information of documents to enhance the representation learning of queries. Some recent query classification models, such as HCL4QC~\cite{zhu2023hcl4qc}, SMGCN~\cite{yuan2024semi}, and HQC~\cite{he2024hierarchical} also explore utilizing the hierarchical category tree structure or instance hierarchy to facilitate models to learn external information beyond query information. 

Industrial methods for query classification typically rely on users’ click behavior to generate training samples. While using real user interactions can improve model accuracy, it also introduces a dependency cycle known as the “Matthew effect.” This cycle leads to biased training data, where popular queries receive excessive focus, skewing the model’s understanding and limiting its ability to generalize to tail queries. Moreover, existing models often handle subtasks separately, overlooking potential synergies that could enhance efficiency in model optimization and development. The lack of a unified framework further impedes the sharing of insights and improvements across different subtasks, thereby restricting overall performance.

To address these challenges, we propose a semi-supervised scalable unified framework (SSUF) for e-commerce query classification. SSUF is designed to overcome the above problems by introducing a set of scalable modules: (1) Label-enhanced module, (2) Knowledge-enhanced module, and (3) Structure-enhanced module to enhance query and label representations with prior knowledge, reduce dependency on posterior labels and enhances the model’s ability to generalize from limited data. Each module within SSUF is designed to be highly pluggable, allowing for flexible adaptation to the specific needs of different subtasks. This modularity ensures that the framework can be tailored to enhance various aspects of query classification.

The contributions of this paper are as follows:
\begin{itemize}
\item We propose a novel unified framework to improve the optimization efficiency of e-commerce query classification models.

\item We design three scalable modules that enhance the query and label representations and break the ``Matthew vicious cycle'' to improve the performance of query classification.

\item We conduct extensive offline and online A/B experiments, and SSUF significantly outperforms existing strong baselines. It has been deployed at an e-commerce platform and brings great commercial value. 
\end{itemize}

\section{Related Work}

\subsection{Multi-label Classification}
Multi-label classification is a vital area in machine learning, where each instance can be linked to multiple labels. Machine learning methods address this problem by transforming the multi-label problem into several single-label tasks~\cite{tsoumakas2007random,tsoumakas2009mining,read2011classifier}. Recently, deep learning models, such as XML-CNN~\cite{liu2017deep}, LSAN~\cite{xiao2019label} and LEAM~\cite{wang2018joint} utilize contextual information or label-specific attention to enhance the interaction between document and labels for classification.

\begin{figure*}[!htbp]
    \centering
    \includegraphics[scale=0.6]{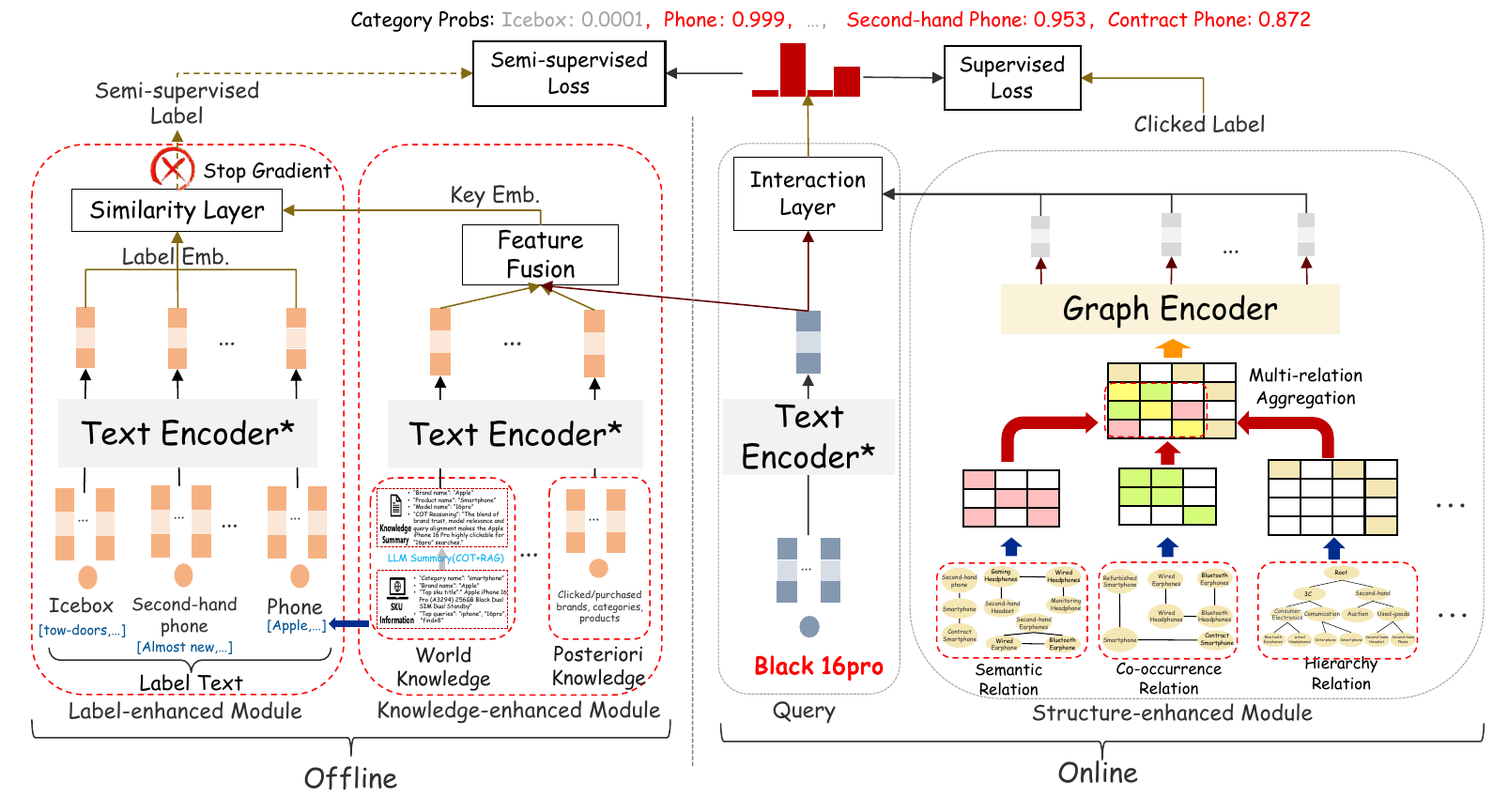}
    \caption{Semi-supervised Scalable Unified Framework for E-commerce Query Classification. The offline part participates in the training of the model but is not directly deployed online. The part with red dashed lines is a pluggable module. The ``Text Encoder*'' denotes a shared text encoder.}
    \label{model_structure}
\end{figure*}

\subsection{Query Classification}
Early models mainly relied on deep learning models, such as CNN~\cite{hashemi2016query}, LSTMs~\cite{sreelakshmi2018deep}, and attention-based models~\cite{zhang2021modeling,yuan2023multi} to extract fine-grained features for classification. Recent works like PHC~\cite{zhang2019improving} explore multi-task frameworks to jointly optimize query classification and textual similarity, while DPHA~\cite{zhao2019dynamic} leverages label graph-based neural networks to model label correlations. HCL4QC~\cite{zhu2023hcl4qc}, SMGCN~\cite{yuan2024semi}, and HQC~\cite{he2024hierarchical} use hierarchical structures and instance hierarchy to learn information beyond query text.

\section{Model}
In this section, we first formally define the query classification task. Then, we describe different modules of SSUF and analyze the influence of the model during the training and inference process.

\subsection{Label-enhanced Module}
Instead of directly using the label index as label embedding, we employ BERT~\cite{kenton2019bert} as the encoder for labels to learn the semantic representation of the label.

The input of the text encoder is a character sequence of label, which is comprised of two parts: (1) the label name $n = [n_1, n_2, \ldots, n_L]$, and (2) the enhanced label side information $m = [m_1, m_2, \ldots, m_{L_m}]$, which is retrieved from (1) label description, such as product words, frequently searched query terms, etc.  (2) world knowledge generated by LLM.  

The label's character sequences are fed into BERT to encode label representation:
\begin{equation}
\mathbf{C}_j = \mathbf{BERT_{CLS}}([n_1, \ldots, n_L, m_1, \ldots, m_{L_m}]) \,, 
\end{equation}
where $\mathbf{C}_j \in \mathbb{R}^{1 \times d}$ is the ``CLS'' representation of the last layer of BERT. In the same way, we can get the representation of query $\mathbf{Q}_i \in \mathbb{R}^{1 \times d}$.

\subsection{Knowledge-enhanced Module}
Industrial methods for query classification have relied on users’ posterior click behavior to generate training samples. However, it leads to the "Matthew vicious cycle" and results in biased training data, where popular queries receive more attention, skewing the model’s understanding and limiting its ability to generalize to less frequent queries. 

We propose a semi-supervised module to overcome the limitations of posterior labels. However, we found that for queries with ambiguous semantics, it is often inaccurate to directly compute semi-supervised labels for queries and labels. For example, the query ``Black 16pro'' refers to an Apple mobile phone model, but due to insufficient semantic information, similarity scores with relevant labels such as ``mobile phone'' and ``second-hand mobile phone'' are low. This results in the semi-supervised signal failing to effectively recall related labels. To solve this issue, we incorporate a knowledge-enhanced module to improve the representation of queries for semi-supervised labeling.

We can use (1) the posteriori knowledge, such as the user's frequently clicked or bought product labels, (2) the world knowledge extracted from LLM as the input. To obtain the world knowledge of the query, we feed the query and the related products to an open-source LLM to summarize a brief description, which may contain relevant queries, categories, products, etc. With this information, the model can comprehensively encode the semantic representations of the query.

After obtaining the posterior and world knowledge, we feed them into a shared text encoder:
\begin{equation}
\mathbf{k}_i = \mathbf{BERT_{CLS}}([k_1, \ldots, k_n ]) \,,
\end{equation}
 to get the knowledge embeddings $\mathbf{K} \in \mathbb{R}^{|K| \times d} $.

To fuse these knowledge embeddings with query representation $\mathbf{Q}_i$, we use an attention module, which can be formulated as follows:
\begin{equation}
\begin{split}
& \mathbf{\alpha} = \mathbf{softmax}(\mathbf{Q}_i  \mathbf{K}^T)  \,,   \\
& \mathbf{q}_i^{\prime} = \mathbf{Q}_i  + \sum_{j=1}^{|K|} \mathbf{\alpha}_j \mathbf{K}_j   \,,   \\
\end{split}
\end{equation}
where $\mathbf{\alpha}$ is the attention score and $\mathbf{q}_i^{\prime} \in \mathbb{R}^{1 \times d}$ is the final fused query representation.

We compute the similarity score between the fused query and label representations to treat it as a semi-supervised label. Specifically, 
\begin{equation}
\begin{split}
& \mathbf{s}_i = \mathbf{stop\_grad} \left( \frac{\mathbf{q}_i^{\prime} \mathbf{C}^T}{\lVert \mathbf{q}^{\prime}\rVert   \lVert \mathbf{C} \rVert} \right)    \,,  \\
& \mathbf{y}^{semi}_{ij} = \mathbf{s}_{ij} \cdot \mathds{1}_{\mathbf{s}_{ij}  \geq \tau}  \,, \\
\end{split}
\end{equation}
where $\mathbf{s}_i \in \mathbb{R}^{1 \times |C|}$ is the relevance scores between query $q_i$ and all categories. $\tau$ is the threshold to filter the categories with low scores. $\mathbf{y}^{semi}_{ij}$ is the semi-supervised label.

Both queries and labels utilize the same text encoder, but their word distributions is different. Feeding the gradient of the semi-supervised signal back into the semi-supervised label module can create a circular dependency, potentially causing the model to collapse. To prevent this, we disable gradient feedback from this branch.

\subsection{Structure-enhanced Module}

\subsubsection{Graph Construction}
Firstly, we obtain the co-occurrence relations between categories by counting the co-occurrence times of categories in the training samples. Then, we compute the conditional probability of two categories and obtain the adjacency matrix $\mathbf{A}^{coo}$: 
\begin{equation}
\begin{split}
& \mathbf{a}_{ij} =\frac{N(c_i, c_j)}{N(c_i)}  \,,   \mathbf{A}^{coo}_{ij} = \mathbf{a}_{ij} \cdot \mathds{1}_{\mathbf{a}_{ij} \geq \alpha}
\end{split}
\end{equation}
where $N(c_i, c_j)$ is co-occurrence frequency of label $c_i$ and $c_j$ and $N(c_i)$ denotes the frequency of label $c_i$. $\alpha$ is the threshold to filter the edges with low relevance scores. $\mathbf{A}^{coo} \in \mathbb{R}^{|C| \times |C|}$ is the adjacency matrix of co-occurrence.

Then, we can obtain the semantic similarity relations between categories by computing the cosine similarity of every pair of categories:
\begin{equation}
\begin{split}
& \mathbf{a}_{ij} = \frac{\mathbf{C}_i \mathbf{C}^T_j}{\lVert C_i\rVert   \lVert C_j\rVert} \,,    \mathbf{A}^{sim}_{ij} = \mathbf{a}_{ij} \cdot \mathds{1}_{\mathbf{a}_{ij} \geq \beta}  \,, \\
\end{split}
\end{equation}
where $\beta$ is the threshold to filter the edges with low relevance scores. $\mathbf{A}^{sim} \in \mathbb{R}^{|C| \times |C|} $ is the similarity adjacency matrix.

For some query classification subtasks, such as intent or category prediction, there is a hierarchical structure among each level of labels. This structure is beneficial in strengthening the relations among relevant labels and weakening the closeness among irrelevant labels. To use this structure, we encode it into the hierarchy adjacency matrix $\mathbf{A}^{hier} \in \mathbb{R}^{|C'| \times |C'|}$, and the edge is defined as:
\begin{equation}
\begin{split}
& \mathbf{A}^{hier}_{ki} = \mathbf{max} \left(\frac{1}{|Child(k)|}, \frac{m_i}{\sum_{j \in Child(k)} m_j} \right)  \,, \\
\end{split}
\end{equation}
where $Child(k)$ is the child node set of $k$, and $i, j \in Child(k)$. $m_j$ is the frequency of node $j$ being clicked by users in the dataset. $|C'|$ denotes the number of all labels, including the first-level, the second-level, and the leaf labels. $|C|$ denotes the number of leaf labels.

\subsubsection{Graph Fusion and Learning}
In addition to the above three label relationship graphs, each subtask can also increase or decrease the number of label graphs based on its existing input data and business characteristics. 

After obtaining the label correlation matrices, we fuse these correlation matrices and normalize the fused matrix with a normalization method~\cite{kipf2017semi}: 
\begin{equation}
\begin{split}
& \mathbf{A} = \frac{1}{2}(\mathbf{A}^{coo} + \mathbf{A}^{sim}) \rightarrow  \mathbf{A}^{hier}  \,,  \\
& \widehat{\mathbf{A}} = \mathbf{D}^{-\frac{1}{2}} \left(\mathbf{A + I} \right)\mathbf{D}^{-\frac{1}{2}} \,,  \\
\end{split}
\end{equation}
where $\rightarrow$ denotes an assignment symbol. The assignment process is shown in Figure~\ref{model_structure}. $\mathbf{A} \in \mathbb{R}^{ |C'| \times |C'|}$ is the final adjacency matrix. $\mathbf{I}$ is a identity matrix. $\mathbf{D}$ is a diagonal degree matrix with $\mathbf{D}_{ii} = \Sigma _{j}\mathbf{A}_{ij}$. Finally, we use GCN~\cite{kipf2017semi} to learn nodes' representation $\mathbf{H} \in \mathbb{R}^{|C'| \times d}$ from the final adjacency matrix $\mathbf{A}$.

Although the training samples for tail labels are limited, these labels can be readily linked to their associated hot labels through intricate label relationships. Such relationships enable the transfer of gradients from samples with hot labels to those with tail labels, leading to more effective representation training for tail labels and mitigating the limitations of posterior labels.

\subsection{Training and Inference}
In our application scene, we only need to classify a user's input query $\mathbf{q}_i \in \mathbb{R}^{1 \times d}$ to the leaf labels space rather than all labels. Thus, we extract from $\mathbf{H}$ to get leaf labels embedding $\mathbf{H}_l \in \mathbb{R}^{|C| \times d}$. Finally, we use an interaction layer to project the query into label space:
\begin{equation}
    \widehat{\mathbf{y}}_i = \mathbf{sigmoid}(\mathbf{q}_i \mathbf{H}_l^T + \mathbf{b}) \,,
\end{equation}
where $\mathbf{b} \in \mathbb{R}^{1 \times |C|}$ is the bias, and $\widehat{\mathbf{y}}_i  \in \mathbb{R}^{1 \times |C|}$ is the predicted labels of query $q_i$.

To optimize the model with the posteriori and priori labels, we fuse them together as follows:
\begin{equation}
\begin{split}
& \mathbf{y}_i = \mathbf{min} \left(\mathbf{y}_i^{click} + \mathbf{y}_i^{semi}, 1.0 \right)  \,,  
\end{split}
\end{equation}
where $\mathbf{y}_i^{click}$ is the multi-hot encoding of clicked labels of query $q_i$, and the value range of $\mathbf{y}_{i} $ is $\mathbf{y}_{i} \in [0, 1]$. We use the binary cross-entropy loss as the objective to train the model.

\section{Experiment}

\subsection{Dataset}
\label{sec:Dataset}
To evaluate the effectiveness of SSUF, we conducted a series of experiments on two large-scale real-world datasets derived from user click logs on an e-commerce application. The statistics of the datasets are listed in Table~\ref{tab:dataset1} and~\ref{tab:dataset2}. The experiments focused on the following two tasks: 
\begin{itemize}
    \item \textbf{Intent Task}: This task predicts multiple purchase intents based on the user’s query. The e-commerce platform meticulously defines a hierarchical intent architecture by experts, encompassing over 1000 distinct user intents. Both the train and test data are extracted from historical user click logs. 
    \item \textbf{Category Task}: This task aims to predict the product categories the user demands. The high-click categories (top 95\% click-through rates) of products previously were considered the query's categories. 
\end{itemize}

\label{sec:appendix1}
\begin{table}[!htbp]
    \centering
    \caption{Data statistics on the intent classification task.}
    \setlength{\tabcolsep}{1.5mm}{
        \begin{tabular}{c|ccc}
            \toprule
            \multirow{2}{*}{\textbf{Statistics}} & \multicolumn{3}{c}{\textbf{Intent Task}} \\ 
            &\textbf{Train} &\textbf{Val} &\textbf{Test}   \\   
            \hline
            \hline 
            Queries  & 67,450,702 & 20,0000  & 31,792 \\
            \hline
            Avg. chars & 7.63 & 5.00   &  8.36   \\
            \hline
            Total Labels  & 1,605 & 1,605    & 1,605     \\
            \hline
            Avg. \# of labels  & 1.04  & 1.67 &  1.91     \\
            \hline
            Min. \# of labels & 1 & 1   &  1   \\
            \hline
            Max. \# of labels & 7 & 3  &  16  \\
            \bottomrule
        \end{tabular}
    }
    \label{tab:dataset1}
\end{table}

\begin{table}[!htbp]
    \centering
    \caption{Data statistics on the category task.}
    \setlength{\tabcolsep}{1mm}{
        \begin{tabular}{c|ccc}
            \toprule
            \multirow{2}{*}{\textbf{Statistics}} & \multicolumn{3}{c}{\textbf{Category Task}}  \\ 
            &\textbf{Train} &\textbf{Val} &\textbf{Test}  \\   
            \hline
            \hline 
            Queries   & 113,686,150 & 20,0000 & 33,960 \\
            \hline
            Avg. chars  & 8.50 & 6.53   &  6.02 \\
            \hline
            Total Labels    & 6,634 & 6,634   &  6,634 \\
            \hline
            Avg. \# of labels   & 1.52 & 2.05   &  5.33  \\
            \hline
            Min. \# of labels  & 1 & 1   &  1  \\
            \hline
            Max. \# of labels & 16 & 13  &  20 \\
            \bottomrule
        \end{tabular}
    }
    \label{tab:dataset2}
\end{table}

\begin{table*}[!htbp]
  \caption{
    The experimental results are compared to multi-label text classification and query classification models. 
  }
  \centering
  \label{tab:experiment}
  \setlength{\tabcolsep}{1.2mm}{
      \begin{tabular}{c|ccc|ccc|ccc|ccc}
                \toprule
                
                \multirow{3}{*}{\textbf{Models}}  & 
                \multicolumn{6}{c|}{\textbf{Intent Task}} & \multicolumn{6}{c}{\textbf{Category Task}} \\
                
                & \multicolumn{3}{c|}{\textbf{Micro}}  
                & \multicolumn{3}{c|}{\textbf{Macro}} 
                & \multicolumn{3}{c|}{\textbf{Micro}}  
                & \multicolumn{3}{c}{\textbf{Macro}}  \\
                
                &\textbf{Prec.} &\textbf{Recall} &\textbf{F1}
                &\textbf{Prec.} &\textbf{Recall} &\textbf{F1}
                &\textbf{Prec.} &\textbf{Recall} &\textbf{F1}
                &\textbf{Prec.} &\textbf{Recall} &\textbf{F1} \\
                \midrule
                \midrule
                XML-CNN   & 78.66 & 32.09 & 45.58 & 50.33 & 20.76 & 27.24  & 86.95 & 24.60 & 38.34  & 40.50 & 15.44 & 20.16  \\
                LEAM      & 76.22 & 37.21 & 50.01 & 55.11 & 25.72 & 32.40  & 76.79 & 26.68 & 39.60  & 39.40 & 17.19 & 21.31  \\
                LSAN      & 76.46 & 34.96 & 47.98 & 54.47 & 25.12 & 31.71  & 86.39 & 23.66 & 37.15  & 44.69 & 17.79 & 22.84  \\
                
                \midrule 
                DPHA      & 77.22 & 36.91 & 49.94 & 55.09 & 25.74 & 32.53  & \textbf{87.29} & 22.49 & 35.76  & 36.08 & 13.11 & 17.26  \\
                MMAN      & \textbf{79.26} & 38.96 & 52.24 & 56.27 & 26.32 & 33.36  & 82.05  &  32.57  &  46.63  & \textbf{57.41}  &  28.26   & 34.68  \\
                HCL4QC    & 74.28 & 40.25 & 52.21 & 54.13 & 31.33 & 37.94 & 79.39 & 33.02 & 46.64 & 54.03 & 30.17 & 36.11 \\
                SMGCN      & 75.83 & 49.91  & 59.72  & \textbf{63.18}  & 43.90  & 48.54   & 82.51 & 40.05 & 53.92  & 55.83 & 35.62 & 40.15 \\
                HQC      & 75.02 & 37.03 & 49.58 & 50.28 & 30.87 & 36.77 & 80.87 & 31.03 & 44.85 & 54.73 & 28.74 & 33.98  \\
                \midrule  
                \textbf{SSUF} & 74.89 & \textbf{52.62} & \textbf{61.81} & 62.74 & \textbf{45.91} & \textbf{49.46} & 80.74 & \textbf{43.40} & \textbf{56.45} & 54.98 & \textbf{36.02} & \textbf{41.22} \\
                w/o. SE-S & 73.49 & 50.92 & 60.16 & 59.49 & 41.32 & 45.21 & 79.92 & 41.31 & 54.47 & 54.36 & 34.34 & 39.72 \\
                w/o. SE-C & 74.03 & 51.19 & 60.53 & 59.92 & 40.21 & 44.92 & 79.17 & 40.91 & 53.94 & 54.12 & 34.92 & 39.24 \\
                w/o. SE-H & 74.32 & 52.02 & 61.20 & 60.33 & 44.02 & 47.29 & 79.32 & 41.88 & 54.82 & 54.43 & 35.13 & 39.95 \\
                w/o. SE       & 76.88 & 48.28  & 59.31  & 56.88  & 37.58  & 43.30   & 81.44 & 38.92 & 52.67  & 55.42 & 34.39 & 38.52 \\
                w/o. KE       & 74.91 & 49.12  & 59.33  & 56.91  & 42.12  & 45.82   & 81.83 & 39.12 & 52.93  & 55.88 & 35.43 & 39.24 \\
                w/o. LE\&KE       & 77.03 & 45.05  & 56.85  & 55.49  & 32.21  & 42.36   & 82.02 & 35.35 & 49.41  & 56.02 & 30.51 & 36.47 \\
                BERT          & 81.28 & 37.59  & 51.41  & 51.63  & 29.97  & 36.84  & 82.83  &  31.99  &  46.15  & 56.72  &  27.80   & 33.80  \\
                \bottomrule
        \end{tabular}
    }
\end{table*}

\subsection{Baseline Models}
We compare SSUF with several strong baselines, including multi-label classification methods and query classification models. The detailed introductions are listed as follows:

\noindent (1) Multi-label classification baselines: 
\begin{itemize}
    \item \textbf{XML-CNN}~\cite{liu2017deep}: It is a CNN-based model, which combines the strengths of CNN models and goes beyond the multi-label co-occurrence patterns. 
    \item \textbf{LEAM}~\cite{wang2018joint}: It is a label-embedding attentive model, which embeds the words and labels in the same space, and measures the compatibility of word-label pairs. 
    \item \textbf{LSAN}~\cite{xiao2019label}: It is a label-specific attention network that uses document and label text to learn the label-specific document representation with the self- and label-attention mechanisms.
\end{itemize}

\noindent (2) Query classification baselines:  
\begin{itemize}    
    \item \textbf{DPHA}~\cite{zhao2019dynamic}: It contains a label graph-based neural network and soft training with correlation-based label representation. 
    \item \textbf{MMAN}~\cite{yuan2023multi}: It is a BERT-based model that extracts features from the character and semantic level from a query-category interaction matrix to mitigate the gap in the expression between informal queries and categories.
    \item \textbf{HCL4QC}~\cite{zhu2023hcl4qc} uses hierarchical structures and instance hierarchy to learn information beyond the query text. 
    \item \textbf{SMGCN}~\cite{yuan2024semi}: It extends category information and leverages categories' co-occurrence and semantic similarity graph to enhance the relations among labels.
    \item \textbf{HQC}~\cite{he2024hierarchical}: It uses hierarchical information by enhanced representation learning that utilizes the contrastive loss to discern fine-grained instance relations in the hierarchy, and a nuanced hierarchical classification loss that attends to the intrinsic label taxonomy. 
\end{itemize}

\subsection{Experiment Settings}
Query classification is essentially a text classification task. In alignment with previous studies~\cite{zhang2021modeling,yuan2023multi}, we evaluate model performance using micro and macro precision, recall, and F1-score metrics. 

Our models are implemented using the PyTorch framework, and we use the Adam algorithm~\cite{kingma2014adam} with learning rate $1e^{-4}$. The BERT embeddings have a dimensionality of 768. We use a 2-layer GCN to learn label embeddings from the graph, with an embedding dimensionality of 768. The maximum query length is set to 20. Edge thresholds ($\alpha$) and ($\beta$) are both set to 0.5, determined by grid search. Model training use a warm start strategy, with the semi-supervised threshold ($\tau$) initially set at 1.0 and gradually decreased to 0.8 during training. Training is conducted over 20 epochs, with a batch size of 1024.

\subsection{Offline Evaluation}

\subsubsection{Offline performance}
The experimental results are shown in Table~\ref{tab:experiment}. Specifically, we have the following observations:
(1) SSUF shows significant performance advantages in both tasks over the multi-label baselines. Although improving query and label representations can alleviate the problem of insufficient contextual information caused by short queries, they ignore the complexity in industrial applications. Industrial datasets suffer from class imbalance, with data distribution heavily skewed towards popular labels, leading to the ``Matthew vicious cycle''. Therefore, the effectiveness of these models may be reduced if directly applied to online systems.

(2) Compared to query classification methods, SSUF also achieves better performance on both tasks. As the results are shown in the table, the recall of relevant categories obtains nearly 3\% F1 improvement on both tasks. Although HCL4QC and HQC also use hierarchical structures to enhance label representations, they cannot model complete label relationships and a priori knowledge to break the vicious cycle. Furthermore, when the query lacks sufficient semantic information, the model's generalization ability is insufficient, and it degenerates into a memory model. SSUF can solve these problems with three extensible modules by fusing posterior signal and a priori knowledge, resulting in superior performance.

\subsubsection{Ablation study}
To discover the relative importance of each module in SSUF, we performed ablation studies on its variants: 

\begin{itemize}
    \item \textbf{w/o KE}:  Removing the knowledge-enhanced module.
    \item \textbf{w/o KE+LE}: Removing the label-enhanced module and knowledge-enhanced module.
    \item \textbf{w/o KE}: Removing the structure-enhanced module.
    \item \textbf{w/o KE-S}: Removing the semantic relation of the structure-enhanced module.
    \item \textbf{w/o KE-C}: Removing the co-occurrence relation of the structure-enhanced module.
    \item \textbf{w/o KE-H}: Removing the hierarchy relation of the structure-enhanced module.
    \item \textbf{BERT}: Only remaining BERT as text encoder for query classification.
\end{itemize}

The experiment results are shown in Table~\ref{tab:experiment}. The experimental results demonstrate that:

(1) When removing the SE, the performance has a little drop compared with SSUF on both datasets. A similar phenomenon can be seen when removing the co-occurrence graph, showing that the similarity or co-occurrence graph contains extra information that is neglected in the posterior data. 

(2) When we eliminate both similarity and co-occurrence graphs, the performance degrades by more than 5\% compared with the complete SSUF. The results indicate that both graphs play different roles in category representation learning. 

(3) After removing these three modules, we can see that the micro and macro F1 decay by 8\% compared with the complete SSUF. This result further demonstrates that all of these components in SSUF provide complementary information to each other, and are requisite for query classification.

\subsection{Online Evaluation}

\subsubsection{Online Deployment}
To reduce the deployment latency, the text encoder of the SSUF used a 4-layer BERT, which is consistent with the online model. Moreover, we only need to cache the category embeddings $\mathbf{H} \in \mathbb{R}^{|C| \times d}$ produced by GCN rather than directly deploying the  GCN. In this way, we can deploy SSUF without adding any additional computation and latency.

\subsection{Online architecture}
\label{sec:appendix2}
\begin{figure}[H]
\centering
\includegraphics[width=8cm,height=6cm]{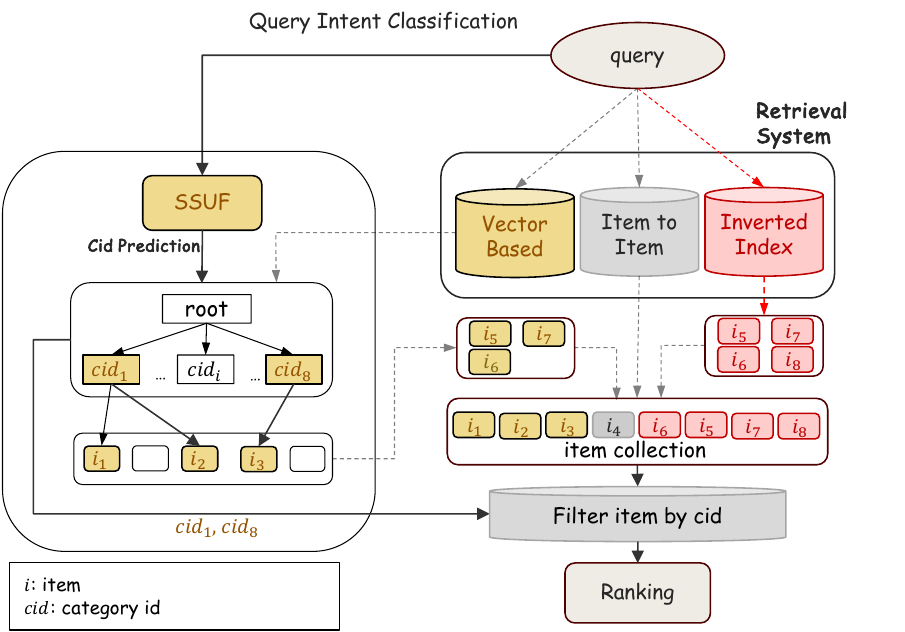}
\caption{The deployment of SSUF and the role of category plays in the E-commerce system.}
\label{fig:system}
\end{figure}

Figure~\ref{fig:system} shows the role of SSUF in the search system. When a user inputs a query, SSUF first predicts the user’s intent and identifies the relevant categories, passing this information to downstream modules. The vector-based retrieval module then finds items associated with these categories. The retrieved items are combined with items from other retrieval sources and filtered by a sub-module to remove those that do not match the user’s desired categories. The filtered items are then sent to the ranking module.

\subsubsection{Online Performance}
We deployed the SSUF and base model in the advertising engine for A/B testing. Each model was allocated 5\% of the traffic. The A/B test was observed for a minimum duration of one week. For online evaluation, we use several business metrics: Imp. (the number of times ads are displayed), Click, CPM (cost per mille), and ad revenue.

\begin{table}[!htbp]
  \centering
  \caption{Online improvements of SSUF. Improvements are statistically significant with $p < 0.05$ on a paired t-test. (\%)}
  \label{online_performance}
  \setlength{\tabcolsep}{1.2mm}{
      \begin{tabular}{c|cccc}
            \toprule
            Models &\textbf{Imp.}  &\textbf{Click} &\textbf{CPM}  & \textbf{Ad. Revenue}
            \\
            \midrule
            Online          & - & -  & - & -   \\
            SSUF        & +3.14   & +2.72  & +1.35    & +4.49   \\
            w/o. SE-S   & +3.07   & +2.38  & +0.90    & +3.97   \\
            w/o. SE-C   & +2.43   & +2.27  & +1.51    & +3.94   \\
            w/o. SE-H   & +2.72   & +2.34  & +1.13    & +3.86   \\
            w/o. SE     & +2.51   & +2.38  & +1.02    & +3.53   \\
            w/o. KE     & +2.67   & +2.34  & +0.95    & +3.61   \\
            w/o. LE     & +2.93   & +2.47  & +1.24    & +4.17   \\
            \bottomrule
        \end{tabular}
    }
\end{table}

As shown in Table~\ref{online_performance}, SSUF achieves significant improvements in business metrics compared to the online model. The improvement of ad impressions and clicks indicates that more relevant products are retrieved by the advertising system, and they are effectively aligned with user preferences and search intentions. The removal of any submodule of SSUF results in a performance decline, which further validates the effectiveness of each module and its synergistic integration within the SSUF. 

In conclusion, both the offline and online experimental results consistently demonstrate the efficiency, universality, and scalability of SSUF.

\section{Conclusion }
In this paper, we propose a semi-supervised scalable unified framework for e-commerce query classification, addressing critical challenges such as short and ambiguous query contexts and the reliance on posterior click behaviors. SSUF integrates three innovative modules: label-enhanced module, knowledge-enhanced module, and structure-enhanced module that collectively improve query and label representations, break the ``Matthew vicious cycle'' and allow for flexible adaptation across different subtasks. Extensive offline and online A/B testing shows that SSUF significantly surpasses baselines, validating its effectiveness and practicality. The successful deployment of SSUF in a commercial e-commerce platform highlights its substantial commercial value. 

In future work, we plan to enhance SSUF by incorporating user-specific information and historical search behaviors to achieve personalized query classification,  aiming to improve classification accuracy and user satisfaction.

\clearpage

\section*{Ethical Consideration}
We discuss the ethical issues from the following aspects:
\begin{itemize}
    \item \textbf{Intended Use.} If the technology operates as intended, both sellers and users of e-commerce platforms can benefit from the SSUF model. SSUF can help customers in quickly identifying the products they desire. It also aids sellers by reducing the effort required to select more accurate product categories when listing new products.
    \item \textbf{Failure Modes.} In the event of a malfunction, SSUF may output inaccurate product information. This non-factual information could potentially influence the shopping experience of users. The system might predict wrong product categories, thereby recommending undesired products to customers and adversely affecting their shopping experience.
\end{itemize}

\bibliography{custom}

\appendix

\end{document}